\documentclass{article}
\usepackage{float}
\usepackage{comment}

\usepackage[preprint]{neurips_2025}

\usepackage[utf8]{inputenc} 
\usepackage[T1]{fontenc}    
\usepackage{hyperref}       
\usepackage{url}            
\usepackage{booktabs}       
\usepackage{amsfonts}       
\usepackage{nicefrac}       
\usepackage{microtype}      
\usepackage{xcolor}         
\usepackage{graphicx}
\usepackage{placeins}

\title{ASCIIBench: Evaluating Language-Model-Based Understanding of Visually-Oriented Text}
\workshoptitle{Multimodal Algorithmic Reasoning}
\author{%
  Kerry Luo \And
  Michael Fu \And
  Joshua Peguero \And
  Husnain Malik \And
  Anvay Patil \AND
  Joyce Lin \And
  Megan Van Overborg \And
  Ryan Sarmiento \And
  Kevin Zhu \AND
  Algoverse AI Research \\
  \texttt{kerryluo1@gmail.com, kevin@algoverseairesearch.org}
}

\begin{document}

\maketitle

\newcommand{\ourmodel}{\texttt{ASCII-Bench}}

\begin{abstract}
Large language models (LLMs) have demonstrated several emergent behaviors with scale, including reasoning and fluency in long-form text generation. However, they continue to struggle with tasks requiring precise spatial and positional reasoning. ASCII art, a symbolic medium where characters encode structure and form, provides a unique probe of this limitation. We introduce ASCIIBench, a novel benchmark for evaluating both the generation and classification of ASCII-text images. ASCIIBench consists of a filtered dataset of 5,135 class-labeled ASCII images and is, to our knowledge, the first publicly available benchmark of its kind. Alongside the dataset, we release weights for a fine-tuned CLIP model adapted to capture ASCII structure, enabling the evaluation of LLM-generated ASCII art. Our analysis shows that cosine similarity over CLIP embeddings fails to separate most ASCII categories, yielding chance-level performance even for low-variance classes. In contrast, classes with high internal mean similarity exhibit clear discriminability, revealing that the bottleneck lies in representation rather than generational variance. These findings position ASCII art as a stress test for multimodal representations and motivate the development of new embedding methods or evaluation metrics tailored to symbolic visual modalities. All resources are available at \url{https://github.com/ASCIIBench/ASCIIBench}.
\end{abstract}

\section{Introduction} \label{introduction}
Scaling language models has been shown to induce emergent capabilities \citep{wei2022emergent}, including those involving positional understanding, such as the generation and editing of \texttt{TikZ} drawings \citep{Bubeck2023SparksOA}.
%
%
We define ASCII art as the intersection of text and vision. While text-based visual systems such as ASCII art have a long history in digital art and design \citep{fogelman2013graphicglitch,fogelman2015graphicglitch}, they have not previously been formalized as benchmarks for studying how language models interpret and reason over visual structure encoded purely in text.
The generation and classification of ASCII art introduces challenges that are distinct from conventional NLP and multimodal benchmarks: characters function as visual primitives rather than semantic tokens, necessitating strict structural regularity seen in other forms of structured data like tables \citep{Chen2022LargeLM}.
In contrast to natural images, ASCII art is both present in the pretraining distribution of unimodal language models and natively aligned with their tokenization schemes, enabling direct evaluation without additional adaptation.

\section{The ASCIIBench Dataset}

We introduce ASCIIBench, a high-quality benchmark for ASCII art understanding and generation. Sourced ethically from \url{ascii.co.uk}, the data underwent a rigorous multi-stage curation pipeline. The final dataset contains 5,135 unique ASCII art pieces across 752 classes (e.g., \texttt{aircraft}, \texttt{birds}). All art is credited to the original creators on \url{ascii.co.uk}. In the absence of explicit licensing, we adhered to standard research practices described in Appendix~\ref{app:data-sourcing}.

\subsection{Data Curation \& Analysis}
Raw ASCII art contains pervasive noise like signatures and tags. An in-depth description of our data cleaning methodology and dataset analysis can be found in Appendix~\ref{app:data-curation} and Appendix~\ref{app:data-analysis}.

\section{Classification}
\paragraph{Models} We evaluated multiple models on classification and generation tasks, including Llama 3-8B, Llama 3-8B-Instruct, GPT-3.5, GPT-4o, GPT-4o-mini, GPT-5-mini, and Claude 3.5 Sonnet, testing text-only, vision-only, and text-vision prompts to compare performance across modalities.

\subsection{Model Testing Procedure} \label{model_testing_procedure}
\begin{figure}[htp]
    \centering
    \includegraphics[width=0.45\textwidth]{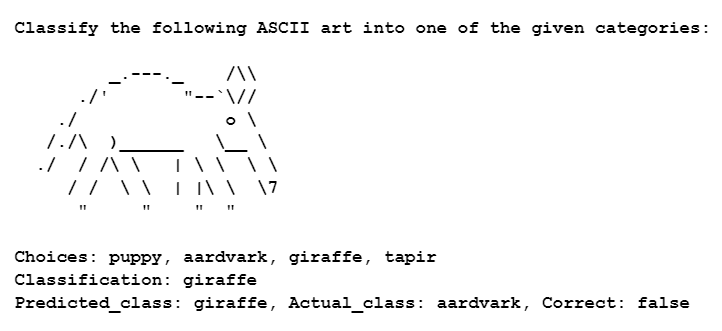}
    \caption{Example classification prompt with result}
    \label{fig:example-classification}
\end{figure}
\paragraph{Prompt}ASCII images are preprocessed based on input modalities. Image preprocessing is described in Appendix~\ref{app:preprocessing}. The model is then prompted to select one of four choices in the format shown in Figure \ref{fig:example-classification}.
\paragraph{Evaluation Metrics}Performance was measured by the model's macro and micro accuracy

\section{Generation}
\paragraph{Models}We prompted GPT-3.5, GPT-4, and GPT-4o to generate 5 ASCII images for each class. 

\paragraph{Approach}To evaluate fidelity, we require an image-to-image metric that captures both the visual and textual characteristics of ASCII art. We use CLIP \citep{radford2021learning}, which aligns images and text through large-scale contrastive training. By comparing embeddings of generated images to reference embeddings derived from ground-truth data, we assess generation accuracy.  

\subsection{Evaluation Metrics}
We leverage CLIP cosine similarity between generated and reference images and representation quality of the embedding space using alignment and uniformity~\citep{wang2022understanding}. We report ROC-AUC for same-class retrieval in Section~\ref{gen_results}. ROC-AUC (Receiver Operating Characteristic – Area Under the Curve) quantifies how well a model separates positive from negative pairs, with 0.5 indicating random performance and 1.0 indicating perfect discrimination.

\subsection{CLIP Cosine Similarity}
\paragraph{Purpose} CLIP cosine similarity is a metric used to evaluate how similar two images are in the context of their high-level features extracted by the CLIP model.
\paragraph{Implementation}ASCII art is rendered following the steps in Appendix~\ref{app:preprocessing} and then embedded with CLIP. The CLIP model, known for its ability to understand high-level visual concepts through natural language supervision, is used to process these images \citep{radford2021learning} . The model extracts feature vectors representing the semantic content of each image.
The cosine similarity score ranges from -1 (completely different) to 1 (exactly the same), with higher scores indicating greater similarity \citep{radford2021learning}.

\subsection{Alignment \& Uniformity}
Alignment measures intra-class compactness, while uniformity quantifies dispersion in the embedding space \citep{wang2022understanding}. Out-of-the-box CLIP shows alignment of $5.85$ (squared $34.20$). Fine-tuning increases alignment to $8.90$ (squared $79.16$) and improves uniformity from baseline to $-7.61$ ($t{=}1$), $-8.09$ ($t{=}5$), and $-8.21$ ($t{=}10$). Together with stable cosine similarities, these results confirm that CLIP is not experiencing representation collapse.

\section{Results}
\subsection{Classification Results}
We evaluate the performance of various models when classifying ASCII art using the methods in Section \ref{model_testing_procedure} with a maximum of 50 output tokens. We report results across three modalities: T (text-only), V (vision-only), and T+V (text+vision). Responses were filtered for possible string parsing errors, resulting in a <2\% average removal. Unfiltered and filtered results are shown in Table~\ref{tab:combined_metrics}.

\begin{table}[H]
\caption{Model performance comparison on raw (left) and filtered (right) datasets.}
\centering
\footnotesize
\setlength{\tabcolsep}{3.5pt} 
\begin{tabular}{@{}c c@{}}
\begin{tabular}[t]{@{}l l c c c@{}}
\toprule
\multicolumn{5}{@{}c@{}}{\textbf{Raw (Unfiltered) Dataset}} \\
\midrule
Model & Mod. & Micro & Macro & Pass \\
      &      & acc. & acc. & rate \\
      &      & (\%) & (\%) & (\%) \\
\midrule
LLaMA3.1-8B-Inst & T & 34.27 & 31.89 & 91.78 \\
LLaMA3.1-8B & T & 29.00 & 25.07 & 82.67 \\
GPT-5-mini & T & 61.60 & 62.39 & \textbf{99.38} \\
 & V & 77.25 & \textbf{84.13} & 99.24 \\
 & T+V & 73.27 & 73.84 & 99.01 \\
GPT-4o-mini & T & 73.61 & 77.60 & 95.23 \\
 & V & 75.72 & 77.77 & 97.27 \\
 & T+V & 76.02 & 77.55 & 96.67 \\
GPT-4o & T & 75.44 & 80.23 & 96.63 \\
 & V & \textbf{77.49} & 82.16 & 98.75 \\
 & T+V & 76.56 & 79.74 & 98.52 \\
GPT-3.5-turbo & T & 39.05 & 33.54 & 91.34 \\
Claude-3.5-Sonnet & T & 59.55 & 56.98 & 98.54 \\
 & V & 76.40 & 76.92 & 99.08 \\
 & T+V & 76.48 & 76.89 & 99.08 \\
\bottomrule
\end{tabular}
&
\begin{tabular}[t]{@{}l l c c c@{}}
\toprule
\multicolumn{5}{@{}c@{}}{\textbf{Filtered Dataset}} \\
\midrule
Model & Mod. & Micro & Macro & Pass \\
      &      & acc. & acc. & rate \\
      &      & (\%) & (\%) & (\%) \\
\midrule
LLaMA3.1-8B-Inst & T & 34.50 & 32.01 & 91.69 \\
LLaMA3.1-8B & T & 29.39 & 25.40 & 82.69 \\
GPT-5-mini & T & 61.36 & 61.97 & \textbf{99.35} \\
 & V & 77.25 & \textbf{84.13} & 99.24 \\
 & T+V & 73.27 & 73.84 & 99.01 \\
GPT-4o-mini & T & 73.52 & 77.27 & 95.19 \\
 & V & 75.72 & 77.77 & 97.27 \\
 & T+V & 76.02 & 77.55 & 96.67 \\
GPT-4o & T & 75.64 & 80.26 & 96.61 \\
 & V & \textbf{77.49} & 82.16 & 98.75 \\
 & T+V & 76.02 & 77.55 & 96.67 \\
GPT-3.5-turbo & T & 39.98 & 33.77 & 91.31 \\
Claude-3.5-Sonnet & T & 59.84 & 57.23 & 98.65 \\
 & V & 76.40 & 76.92 & 99.08 \\
 & T+V & 76.48 & 76.89 & 99.08 \\
\bottomrule
\end{tabular}
\\ \end{tabular}
\label{tab:combined_metrics}
\end{table}

\begin{table}[H]
\caption{Model performance comparison on the filtered dataset.}
\centering
\footnotesize
\setlength{\tabcolsep}{8pt} 
\begin{tabular}{@{}l l c c c@{}}
\toprule
\multicolumn{5}{@{}c@{}}{\textbf{Filtered Dataset}} \\
\midrule
Model & Modality & Micro acc. (\%) & Macro acc. (\%) & Pass rate (\%) \\
\midrule
LLaMA3.1-8B-Inst & Text & 34.50 & 32.01 & 91.69 \\ LLaMA3.1-8B & Text & 29.39 & 25.40 & 82.69 \\ GPT-5-mini & Text & 61.36 & 61.97 & \textbf{99.35} \\ & Vision & 77.25 & \textbf{84.13} & 99.24 \\ & Text+Vision & 73.27 & 73.84 & 99.01 \\ GPT-4o-mini & Text & 73.52 & 77.27 & 95.19 \\ & Vision & 75.72 & 77.77 & 97.27 \\ & Text+Vision & 76.02 & 77.55 & 96.67 \\ GPT-4o & Text & 75.64 & 80.26 & 96.61 \\ & Vision & \textbf{77.49} & 82.16 & 98.75 \\ & Text+Vision & 76.02 & 77.55 & 96.67 \\ GPT-3.5-turbo & Text & 39.98 & 33.77 & 91.31 \\ Claude-3.5-Sonnet & Text & 59.84 & 57.23 & 98.65 \\ & Vision & 76.40 & 76.92 & 99.08 \\ & Text+Vision & 76.48 & 76.89 & 99.08 \\
\bottomrule
\end{tabular}
\label{tab:filtered_metrics}
\end{table}

\subsubsection{Interpretation}
Our results align with those of \citet{jia2024visualperceptiontextstrings}. Larger models had greater performance, and all accuracy values were over 25\%, indicating that models did not choose arbitrarily. Across both raw and filtered datasets, we find that vision-only models consistently outperform text-only and text+vision counterparts, with GPT-4o achieving the highest macro accuracy at 82.2\%. Text-only performance lags significantly, especially for LLaMA and GPT-3.5, underscoring the difficulty of modeling ASCII art as pure text. Surprisingly, adding text to vision does not improve performance and in some cases degrades it, suggesting that current multimodal fusion strategies do not capture ASCII structure effectively. Filtering has little effect on overall trends, indicating robustness of the observed modality gaps.


\subsection{Generation Results}
\label{gen_results}
On unfiltered generations, CLIP showed weak class separation (ROC-AUC $\approx$ 0.55; silhouette $-0.46$), and t-SNE revealed no clear clusters. After filtering inconsistent generations (std $>0.15$, mean similarity $<0.3$), ROC-AUC rose to $0.83$, demonstrating that CLIP can discriminate effectively when ASCII generations are semantically consistent. This indicates that the bottleneck lies in the quality of LLM-generated ASCII rather than in the evaluator.

\begin{figure}[h!]
    \centering
    \includegraphics[width=0.6\linewidth,height=0.35\textheight,keepaspectratio]{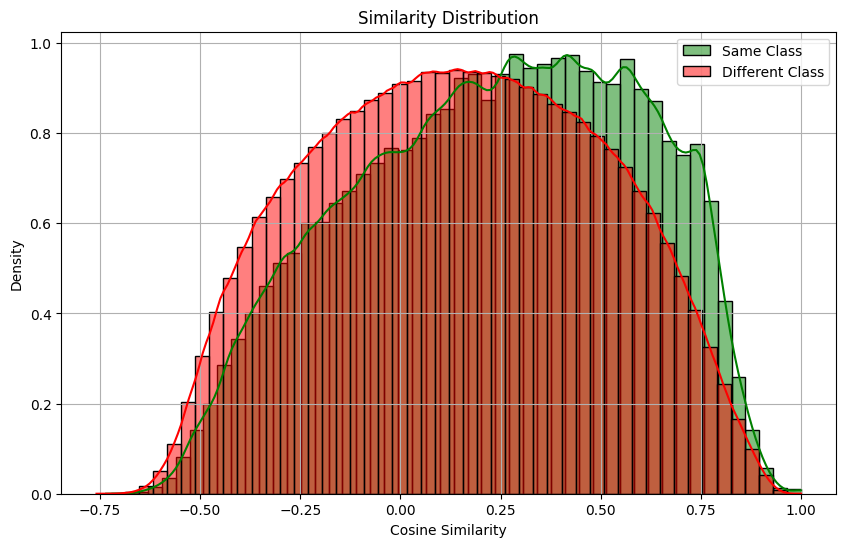}
    \caption{Cosine similarity distributions. Green indicates positive (intra-class) pairs, red indicates negative (inter-class) pairs.}
    \label{fig:similarity-distributions}
\end{figure}

Shown in Figure~\ref{fig:similarity-distributions}, while there is a separation between inter- and intra-class distributions, the separation is not complete, with substantial overlap remaining.

\subsubsection{CLIP Representation Analysis}
We examined cosine similarities, silhouette scores, and t-SNE visualizations of CLIP embeddings (Figure~\ref{fig:TSNE Visualization}). Out-of-the-box CLIP produced weak intra- and inter-class separation (AUC $\approx$ 0.558, silhouette $-0.46$), and fine-tuning with triplet loss yielded only modest gains. Filtering noisy generations did not resolve this, as even the lowest-variance subsets approached near chance performance (AUC = 0.641). However, when restricting analysis to classes with high mean similarity, AUC increased to 0.83, indicating that CLIP can represent ASCII structure only for a subset of well-formed categories. These results show that the primary limitation lies in CLIP’s representational capacity for ASCII art, along with variance in model generations.

\section{Limitations}
Our findings show that evaluation quality depends strongly on input consistency. CLIP performs well only when generations are visually coherent and semantically aligned, but typical LLM outputs are noisy and inconsistent, especially for vague categories. This highlights a dual bottleneck: the instability of ASCII generation and the limited ability of a broad, general-purpose model like CLIP to represent ASCII structure. Filtering demonstrates an upper bound of performance but is not a sustainable evaluation strategy, as it amounts to testing on inputs already close to the training distribution. Future work should explore specialized, smaller models, which may capture ASCII-specific patterns more effectively than CLIP.

\section{Conclusion}
We introduce ASCIIBench, a benchmark for evaluating ASCII art on classification and generation tasks, and used it to probe how multimodal models represent symbolic visual inputs. Empirically, vision-only models consistently outperform text-only and text+vision settings on classification, while CLIP-based evaluation of generations provides limited class separation on unfiltered outputs and improves primarily for classes with high internal similarity. These trends position ASCII art as a stringent stress test for multimodal reasoning: performance hinges on both the consistency of generations and the representational suitability of the embedding model for ASCII structure. Looking ahead, we advocate standardized rendering and preprocessing protocols to enable fair cross-model comparisons, improved prompting and training strategies for ASCII generation, and exploration of structure and variance-aware metrics to better capture and evaluate symbolic layout.

\bibliographystyle{plainnat}   
\bibliography{anthology,custom}





\appendix
\section*{Appendix}   

\FloatBarrier
\section{Data Sourcing}\label{app:data-sourcing}
We express our gratitude to the ASCII artists. We made slight modifications to the original ASCII art and provide the URL to the source of our data. Our dataset is licensed under CC BY NC 4.0.

\section{Data Curation}\label{app:data-curation}
We developed a custom web crawler and an 11-step automated pipeline to remove these artifacts, followed by a multi-stage manual review described in Appendix~\ref{app:noise-removal}. Abstract or ambiguous categories (e.g. "small") were excluded, retaining only well-defined classes. Three annotators then applied a strict rubric to eliminate pieces with: (1) inappropriate content, (2) excessive intra-class variation, (3) overly complex structures, or (4) low quality. This conservative process, requiring strong annotator agreement, removed over 13,000 low-quality images and 1,800 ambiguous classes, resulting in a focused, high-quality benchmark.

\section{Data Analysis}\label{app:data-analysis}
\begin{figure}[h!]
    \centering
    \includegraphics[width=1\linewidth]{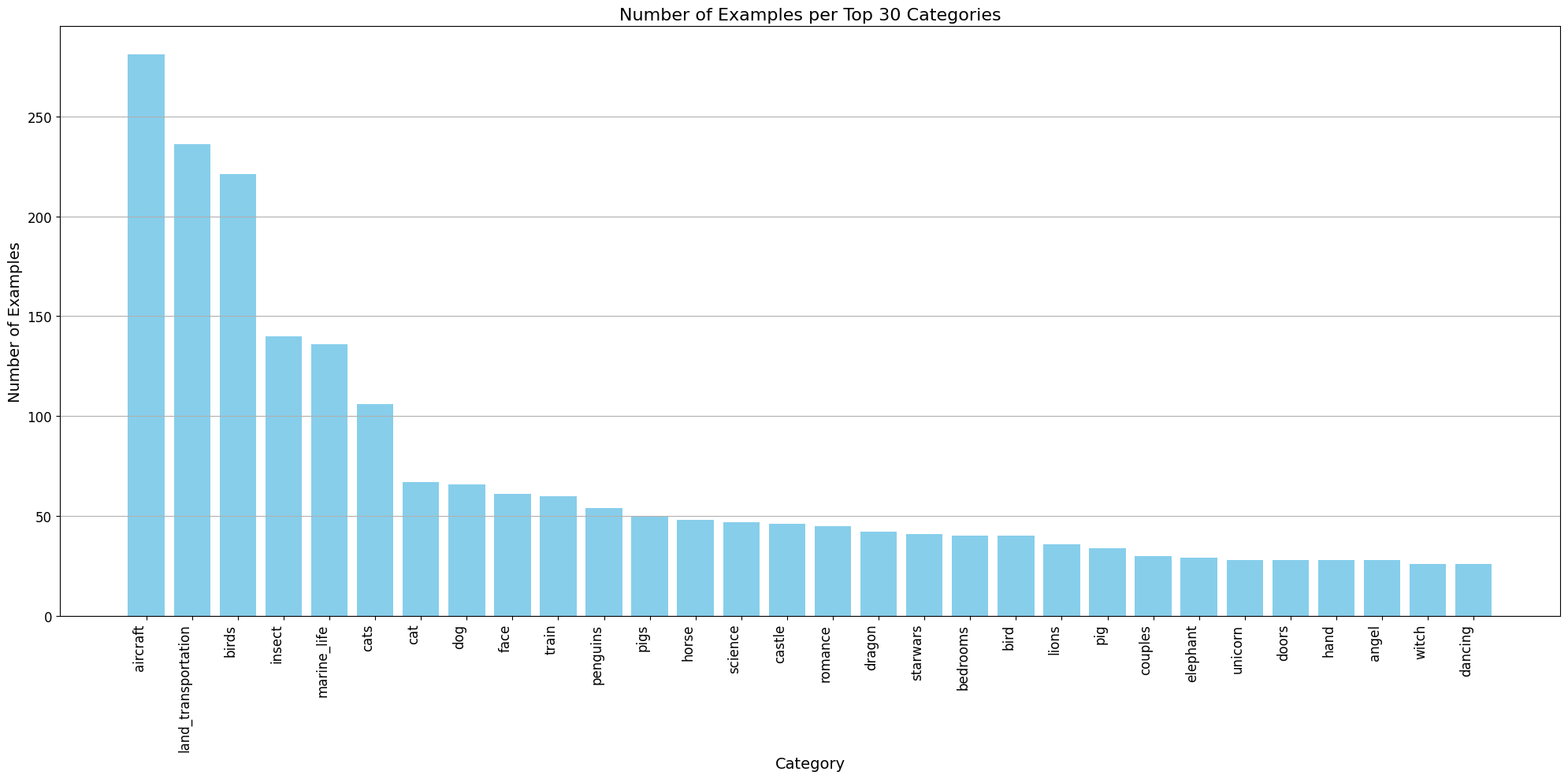}
    \caption{Top 30 Class Histogram}
    \label{fig:Top 30 Class Histogram}
\end{figure}

\begin{figure}[h!]
    \centering
    \includegraphics[width=1\linewidth]{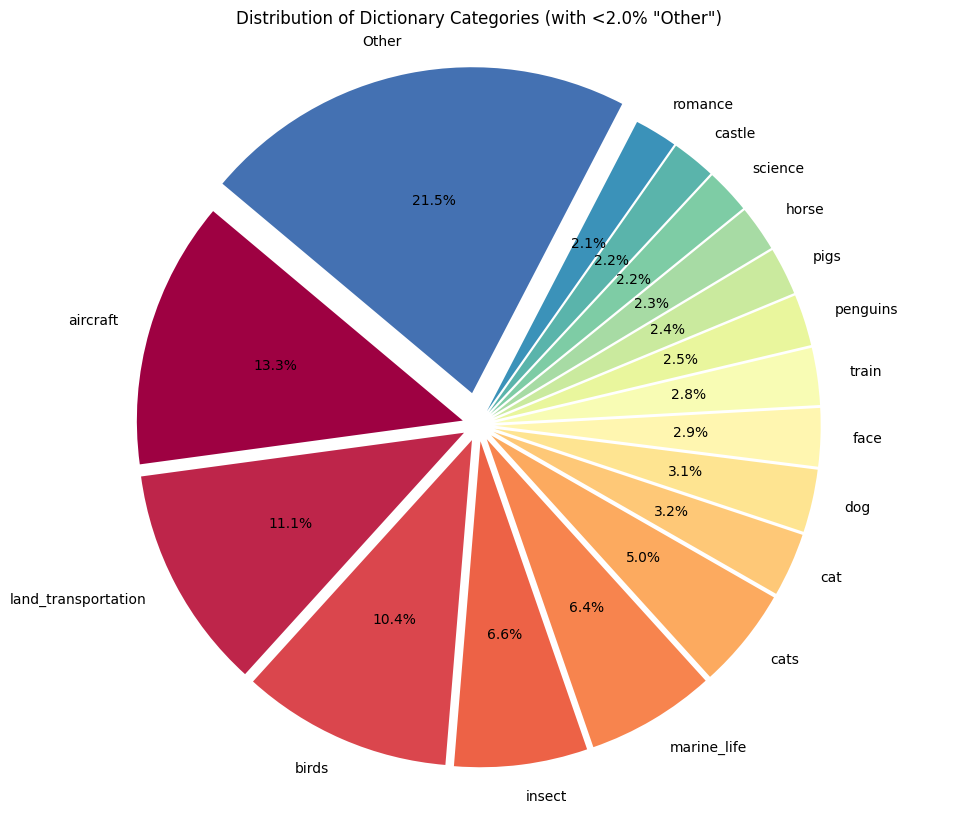}
    \caption{Class Distribution Pie Chart}
    \label{fig:Class Distribution Pie Chart}
\end{figure}

\begin{figure}[h!]
    \centering
    \includegraphics[width=1\linewidth]{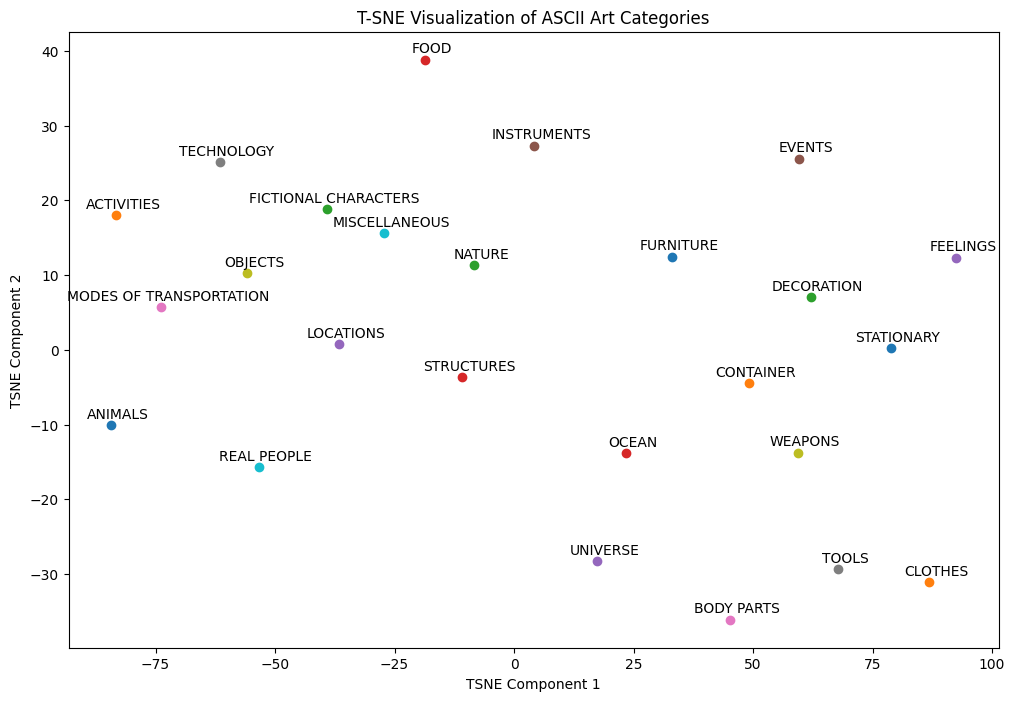}
    \caption{T-SNE visualization of class embeddings}
    \label{fig:TSNE Visualization}
\end{figure}

The curated dataset exhibits a natural long-tail class distribution (Figure~\ref{fig:Top 30 Class Histogram}).
The largest categories are aircraft (13.3\%), land transportation (11.1\%), and birds (10.4\%) (Figure~\ref{fig:Class Distribution Pie Chart}). 
A t-SNE visualization of class embeddings 
(Figure~\ref{fig:TSNE Visualization}) 
confirms semantic coherence, showing clear clustering of related concepts (e.g. animals), demonstrating that ASCII art encodes learnable semantic structures. Character frequency analysis (Figure~\ref{fig:Character Frequency 
Histogram}) 
reveals the artistic "vocabulary": the space character is dominant (>1.6M occurrences), followed by structural elements like \texttt{-}, \texttt{|}, and \texttt{\_}. Alphanumeric characters are used sparingly as accents.

\section{Dataset Analysis Figures}
    \begin{figure}[h!]
    \centering
    \includegraphics[width=0.8\linewidth]{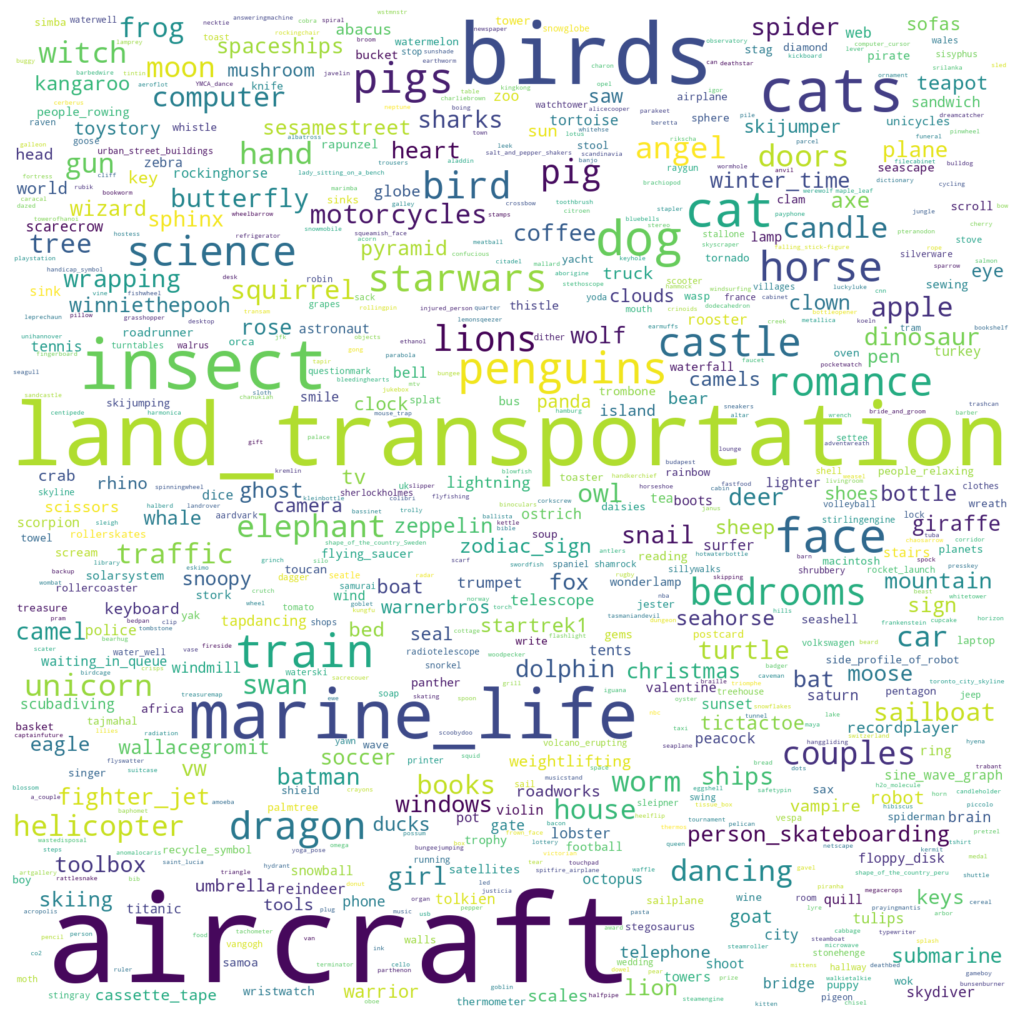}
    \caption{Word cloud of class labels}
    \label{fig:Class Word Cloud}
\end{figure}
\begin{figure}[h!]
    \centering
    \includegraphics[width=1\linewidth]{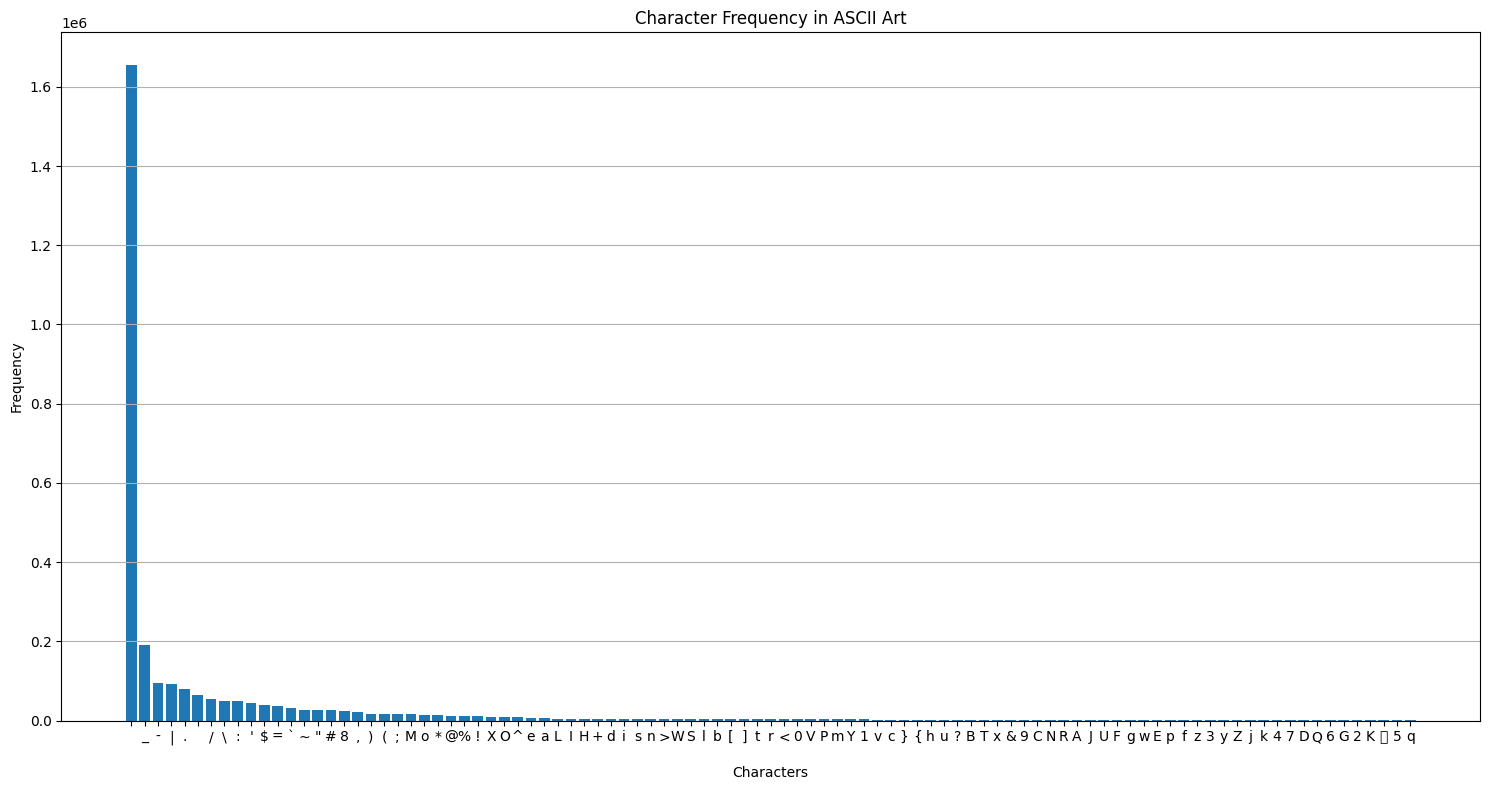}
    \caption{Character Frequency Histogram}
    \label{fig:Character Frequency Histogram}
\end{figure}

\section{Noise Removal Pipeline} \label{app:noise-removal}
\begin{enumerate}
\item Tags consisting of three or fewer alphabetic characters are replaced with whitespace.
\item Creator tags in the last three lines of the artwork are detected and removed while preserving structural spacing.
\item Non-visible Unicode control characters are filtered out.
\item Alphanumeric creator tags appended to the end of the ASCII art are removed.
\item Full names and abbreviated creator tags (e.g., \texttt{'Matthew Kenner'} or \texttt{'jps'}) positioned on the right margin are eliminated.
\item Tags labeled as \texttt{'unknown'} are discarded.
\item Left-aligned creator signatures are identified and removed.
\item Common date formats (e.g., \texttt{12/21/2023}, \texttt{21nov2023}, \texttt{12.21.2023}) are detected via expression matching and subsequently stripped.
\item Contact information, such as email addresses, is localized and pruned.
\item Three-letter creator tags enclosed in dashes (\texttt{--}) or brackets (\texttt{[]}) are filtered out.
\item Known problematic signatures, maintained in a blacklist, are systematically removed.
\end{enumerate}

\section{Preprocessing}\label{app:preprocessing}
We follow \citet{jia2024visualperceptiontextstrings}, using a black monospaced font (DejaVu Sans Mono) on a white background. No blur is added to preserve structural integrity. 

\section{Non-Monospaced Font Ablation}\label{app:ablation}
To examine whether models rely on spatial alignment rather than textual content, we replaced the fixed-width (monospaced) ASCII font with a proportional (non-monospaced) one. As shown in Table~\ref{tab:nonmono}, the results remain nearly unchanged, suggesting that models primarily depend on optical character recognition (OCR)-like mechanisms rather than explicitly reasoning about the positional structure of characters.

\begin{table}[h!]
\centering
\begin{tabular}{lcc}
\toprule
\textbf{Model} & \textbf{Modality} & \textbf{Accuracy} \\
\midrule
GPT-5 & Vision \& Text & 0.7057 \\
GPT-5 & Vision Only & 0.7118 \\
\bottomrule
\end{tabular}
\vspace{2mm}
\caption{Performance on ASCII art rendered in non-monospaced font. The minimal difference suggests reliance on OCR-like recognition rather than positional reasoning.}
\label{tab:nonmono}
\end{table}

\section{Related Works}
\textbf{Emergent Behaviors:} LLMs have been studied for their emergent properties as they scale. \citet{wei2022emergent} highlights that larger models, ones with more parameters and diverse data, possess emergent abilities such as improved reasoning and fluency in text generation. Our work extends this to ASCII-text image generation, which requires textual understanding and visual creativity, skills not typically emphasized in standard LLM evaluations.

\textbf{LLM ASCII Word Recognition:} The Beyond the Imitation Game benchmark (BIG-bench) introduced by \citet{srivastava2023imitation} addresses the need to understand the capabilities of LLMs across various tasks. Our research focused on its ASCII word recognition dataset. Similarly, the ArtPrompt jailbreak attack highlights the need for the improvement of LLM performance in identifying ASCII text, an ability crucial to prevent the circumvention of safeguards and elicitation of unintended behaviors \citep{jiang2024artprompt}. 

\textbf{Other Benchmarks: }ASCII- and Unicode-based visual systems have an established artistic
lineage predating their use in ML \citep{fogelman2013ascii, fogelman2015ascii, fogelman2010museum}. Recent efforts by \citet{jia2024visualperceptiontextstrings} have explored the visual perception capabilities of large language and multimodal models through ASCII art, introducing the \textit{ASCIIEval} benchmark and a corresponding training set to evaluate models’ ability to interpret visual semantics in text strings. Their study establishes ASCII art as a modality-agnostic medium bridging textual and visual understanding, offering a valuable testbed for analyzing perception and modality fusion in modern models. However, while \citet{jia2024visualperceptiontextstrings} constructed the dataset and presented detailed benchmarking results, the accompanying resources were not publicly released at the time of publication. In contrast, our work provides the first publicly available implementation and dataset release for ASCII-based multimodal reasoning, enabling full reproducibility and community-driven extensions of this emerging research direction.

\textbf{Vision-Language Integration and Multimodal Models:} GPT-4V(ision) produces human-aligned scores with detailed explanations, showing promise as a universal automatic evaluator despite some limitations  \citep{zhang2023gpt4vision}. Flamingo models demonstrate strong few-shot learning capabilities, showcasing potential to give LLMs adaptive abilities and decreased dependence on large task-specific datasets \citep{alayrac2022flamingo}. \citet{ramesh2021zeroshot} introduces an autoregressive transformer-based approach for text-to-image generation with competitive zero-shot performance compared to domain-specific models. These capabilities lead into ASCII art generation, which poses unique challenges due to merging textual and visual information. However, the advanced multimodal reasoning of these models also introduces new vulnerabilities.\citep{jia2024visualperceptiontextstrings} demonstrate that Large Vision-Language Models (LVLMs) like LLaVA and GPT-4V are highly susceptible to \textit{Self-Generated Typographic Attacks}, where the model itself is leveraged to create deceptive text and descriptions that cause misclassification, reducing accuracy by up to 60\%. This underscores a critical weakness in how LVLMs fuse and weight visual and textual signals. Prior works have explored the generation of stylized textual outputs using neural networks, with some focusing specifically on the artistic transformations of text and image data \citep{Matsumoto2018ASCIIAC}. Models like ViLBERT and VisualBERT use multimodal pre-training to boost performance in vision-language tasks \citep{li2019visualbert} \citep{lu2019vilbert}. They guide our fine-tuning of the CLIP model for effective ASCII art generation from text descriptions.

\textbf{Evaluation Metrics and Methodologies:} Finally, our evaluation of LLM outputs uses standard metrics used in both language and image processing domains. Metrics such as FID scores, typically used to assess image quality, were adapted to assess the uniqueness and clarity of ASCII-text images produced by our model. We employ CLIP for its ability to effectively bridge text and image representations. Research by \citet{radford2021learning} demonstrates its robust performance in zero-shot classification tasks. This makes CLIP ideal for our needs, as our models must both generate and classify ASCII images from minimal prompts.

\subsection{Other ASCII generation methods}
The exploration of AI in the context of ASCII art has witnessed growing interest in recent years, with researchers exploring methods to optimize conversion accuracy. There are many notable contributions in this field. \citep{goodfellow2014generative} proposed a new framework for estimating generative models via an adversarial process. Researchers have delved into the use of GANs to generate realistic and well-designed ASCII art. By leveraging the adversarial training paradigm, these models can produce a large range of high-quality ASCII representations. Similarly, \citep{gatys2015neural} explored Convolutional Neural Networks and their ability to create artistic imagery by experimenting with the style and content of an image, also known as Neural Style Transfer. \citep{jing2018neural} provides an extension of this idea, comparing different Neural Style Transfer qualitatively and quantitatively. They discuss applications of NST and problems to be addressed in future research. Studies have investigated how deep neural networks can be trained to transfer artistic styles onto ASCII images, showcasing the potential for creative synthesis.

\subsubsection{Transfer Learning}
Naturally, it is also extremely important for the models to efficiently produce accurate results. \citep{9134370} reviews more than forty representative transfer learning approaches from a data and model perspective. Their paper provides more than twenty experiments of different learning models’ performances. The exploration of the efficacy of transfer learning in training models for ASCII art generation and leveraging pre-trained models on large datasets allow researchers to enhance the efficiency and artistic quality of AI-generated ASCII images.

\subsubsection{Methods \& Metrics}
Our research paper took inspiration from methods mentioned in \citep{pang2021imagetoimage}, which explored developments in I2I translations and analyzed key techniques to elaborate on the effect of I2I on the research and industry community. The paper introduced the problem setting of the image-to-image translation task, introduced the generative models used for I2I methods, and discussed the work and applications of multi-domain I2I tasks. Many of these methods and metrics mentioned in this paper are parallel to evaluation metrics we implemented, but while this paper mainly evaluated methods on I2I, our research drew comparisons between these methods and ASCII image generation standards. 

\subsubsection{Frechet Inception Distance}
More specifically, Frechet Inception Distance, which produces an FID score, has been the most widely used metric for measuring the similarity between real and generated images. Muhammad \citep{naeem2020reliable} focuses on the reliability of certain methods. They concluded that while variants of precision and recall metrics are generally unreliable methods, density and coverage metrics will provide more interpretable and reliable comparisons. While precision metrics can overestimate the manifold around real outliers, density fixes this issue by more accurately representing the distribution around real samples. The objective of coverage is to improve upon the recall metric. When models generate many unrealistic and diverse samples, this can skew the data and lead to a false increase in the recall measure. Coverage addresses this by building the manifolds around real samples as opposed to fake ones. This approach is less prone to overestimation since real samples tend to have fewer outliers compared to generated samples. The goal for our purposes would be to capture how well the generated ASCII art (fake samples) represents the original ASCII art (real samples) in both details (density) and overall composition (coverage).

\subsubsection{Interactive System for Structure-based ASCII Art Creation}
Interactive structure-based systems \citep{10.1145/1833349.1778789} invite users to actively participate in the creation of ASCII art; the interactive paradigm empowers users to foster a collaborative synergy between human creativity and computational assistance. While it is an earlier paper, \citep{article} proposes to input images divided into grids for glyph matching using four metrics employed for converting images into ASCII art: template matching which considers pixel positions for dissimilarity measure, normalized cross-correlation which minimizes the influence of line width differences using histograms, Histogram of Oriented Gradients (HOG) representing line directions, and distance transformation indicating line positions. In comparison, our work focuses mainly on AI generation, so while it is not as collaborative it focuses on the optimization and comparison of methods for comparing the output of ASCII art utilizing either tone-based style, which is a detailed and comprehensive image, or structure-based style, which is a simple outline of the image using less characters.

\subsubsection{Perceptual Metrics for Image Quality Assessment}
The recent success of perceptual messages based on deep neural networks in regards to the Image Quality Assessment (IQA) task \citep{kazmierczak2022study} has led to a growing interest in new metrics that outperform previous metrics to develop perceptual information at different resolutions. Whereas the IQA metric is generally easily perceivable for humans, it is more difficult to set a metric for a computational algorithm. Our work investigates the model’s abilities to generate accurate images by comparing it to Euclidean distance and the SSIM index, groundwork laid by \citep{9757209}.

\subsubsection{ASCII Representation Learning} 
While prior work has explored ASCII conversion from images \citep{Matsumoto2018ASCIIAC}, little attention has been given to understanding how models internally represent ASCII structures. Our probing of CLIP embeddings extends this line of inquiry, revealing architecture and how ASCII images are represented in the model.

\end{document}